\documentclass[10pt,twocolumn,letterpaper]{article}

\usepackage{cvpr}              % To produce the CAMERA-READY version
\usepackage{graphicx}
\usepackage{amsmath}
\usepackage{amssymb}
\usepackage{booktabs}
\usepackage{algorithm}
\usepackage{algpseudocode}
\usepackage{physics}
\usepackage{bm}
\usepackage{amsfonts}
\usepackage{bbm}
\usepackage{accsupp}
\usepackage{comment}
\usepackage{makecell}
\usepackage{subcaption}
\usepackage[table,dvipsnames]{xcolor}

\usepackage[pagebackref=false,breaklinks,colorlinks]{hyperref}

\usepackage[capitalize]{cleveref}

\crefname{section}{Sec.}{Secs.}
\Crefname{section}{Section}{Sections}
\Crefname{table}{Table}{Tables}
\crefname{table}{Tab.}{Tabs.}
\colorlet{colorfirst}{Green!25}       
\colorlet{colorsecond}{SpringGreen!65} 
\colorlet{colorthird}{SpringGreen!45}      
\newcommand{\first}{\cellcolor{colorfirst}\bf}
\newcommand{\second}{\cellcolor{colorsecond}}
\newcommand{\third}{\cellcolor{colorthird}}

\def\confName{CVPR}
\def\confYear{2023}
\pdfoutput=1

\begin{document}

\title{Quantum Annealing for Single Image Super-Resolution}

\author{Han Yao Choong \\
ETH Z\"urich\\
Switzerland\\
{\tt\small hchoong@student.ethz.ch}
\and
Suryansh Kumar$\thanks{Corresponding Author (k.sur46@gmail.com)}$\\
ETH Z\"urich\\
Switzerland\\
{\tt\small sukumar@ethz.ch}
\and
Luc Van Gool\\
ETH Z\"urich\\
Switzerland\\
{\tt\small vangool@ethz.ch}
}
\maketitle

%%%%%%%%% ABSTRACT
\begin{abstract}
This paper proposes a quantum computing-based algorithm to solve the single image super-resolution (SISR) problem. One of the well-known classical approaches for SISR relies on the well-established patch-wise sparse modeling of the problem. Yet, this field's current state of affairs is that deep neural networks (DNNs) have demonstrated far superior results than traditional approaches. Nevertheless, quantum computing is expected to become increasingly prominent for machine learning problems soon. As a result, in this work, we take the privilege to perform an early exploration of applying a quantum computing algorithm to this important image enhancement problem, i.e., SISR. Among the two paradigms of quantum computing, namely universal gate quantum computing and adiabatic quantum computing (AQC), the latter has been successfully applied to practical computer vision problems, in which quantum parallelism has been exploited to solve combinatorial optimization efficiently. This work demonstrates formulating quantum SISR as a sparse coding optimization problem, which is solved using quantum annealers accessed via the D-Wave Leap platform. The proposed AQC-based algorithm is demonstrated to achieve improved speed-up over a classical analog while maintaining comparable SISR accuracy\footnote{Due to the limited availability of quantum computing resources for academic research, it is difficult to \textbf{fully} demonstrate and document the true potential of quantum computing on the inference time and accuracy.}.
\end{abstract}

\section{Introduction}\label{sec:intro}

The problem of single image super-resolution (SISR) aims at the reconstruction of a credible and visually adequate high-resolution (HR) image from its low-resolution (LR) image representation. Unfortunately, this problem is inherently ill-posed, as each independent pixel could be mapped to many HR image pixels depending upon the given upsampling factor. Nevertheless, a reasonable solution to SISR is critical to many real-world applications concerning image data, such as remote sensing, forensics, medical imaging, and many more. One of the most attractive properties of SISR for these applications is that it can generate desirable HR images that may not be realizable by onboard camera hardware.

\begin{figure}[t]
\begin{center}
\includegraphics[width=1.0\columnwidth]{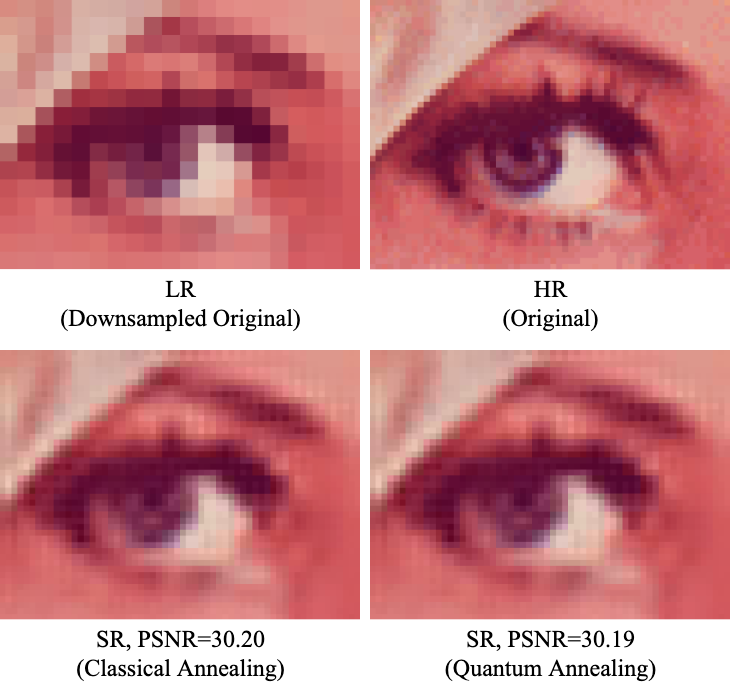}
\end{center}
\caption{Results on a cropped region of the Lenna image with a magnification factor of 3. \textbf{Top row.} Ground-truth LR, HR. \textbf{Bottom row.} \textit{Left:} Result of the Classical Annealing approach showing PSNR value of 30.20  with an inference time of 658.36s. \textit{Right:} Result of our Quantum Annealing approach, showing similar PSNR accuracy of 30.19 with an inference time of \textcolor{red}{89.60}s.}
\label{fig:lena_eye1}
\end{figure}

In recent years, SISR methods based on deep learning (DL) have become prominent, demonstrating DL's suitability on different metrics such as PSNR accuracy, inference speed, memory footprint, and generalization \cite{lai2017deep, yang2020learning, fuoli2020aim, yang2022ntire, wu2021trilevel}. In this paper, our goal is not to argue against the prevailing view on the performance that could be achieved on the available computing machines via deep-networks but to explore the suitability of quantum computing for SISR. Such a motivation seems possible now since very recently functional quantum computers are available for experimentation and research---although in a limited capacity \cite{mcgeoch2019practical, adedoyin2018quantum, neukart2017traffic}. At present, in the domain of quantum computing, quantum processing units (QPUs) carrying out adiabatic quantum computing (AQC) have reached a critical phase in terms of the number of qubits per QPU, making application to practical computer vision problems a reality \cite{benkner2021q, golyanik2020quantum}.

The margins of this draft are limited to discuss all the advantages of quantum computing, and readers may refer to \cite{shor1994algorithms, simon1997power} for deeper insights. The motivation of our work is to investigate the use of current adiabatic quantum computers and the benefits they can have compared to a well-known optimization formulation in SISR. That brings us to formulate the quantum SISR problem as quadratic unconstrained binary optimization (QUBO) problem, posing a challenge to map it to the quantum computer. Specifically, our approach relies on a popular assumption in sparse coding \ie, a high-resolution image can be represented as a sparse linear combination of atoms from a dictionary of high-resolution patches sampled from training images \cite{yang2008image}.

Consequently, our contribution is demonstrating a practical SISR approach using AQC. To this end, a QUBO formulation for sparse coding is developed and adapted for D-Wave Advantage quantum annealers. The resulting algorithm can carry out approximate sparse coding and is benchmarked against a classical sparse coding approach \cite{yang2008image} and a state-of-the-art deep learning SISR model \cite{liang2021swinir}. As an early work in the synthesis of AQC for SISR, the approach proposed in this paper can serve as a basis for the further development of algorithms to exploit quantum parallelism better. As mentioned before, our proposed AQC algorithm achieves improved speed-up over a classical approach with similar accuracy when tested on popular datasets.

\section{Related Work}\label{sec:rw}
Since our approach is one of the early approaches for solving SISR on quantum computing approaches, we discuss some popular SISR methods and existing quantum computing-based computer vision methods in this section.

\smallskip
\noindent
\textbf{\textit{(a)} Single Image Super-Resolution (SISR).}
As is well-known,  HR images are related to their LR counterparts by a non-analytic downsampling operation with generally permanent information loss \cite{timofte2015a+}. As such, with HR images being the causal source of observed LR images, the inference of HR from LR images casts SISR to be an inverse problem  that approximates the inverse of the non-analytic downsampling, with the possibility of utilizing assumptions or prior knowledge about the HR-to-LR transformation model \cite{kathiravan2014overview, yang2008image}. Furthermore, SISR has the property of being ill-posed, as a single LR image observation can have multiple HR mapping as valid non-unique solutions \cite{kathiravan2014overview}.

For decades, SISR has been an active research area with different approaches to solve it reliably and efficiently \cite{pasztor2000learning, timofte2015a+, yang2008image}. Among early approaches are observation model-based methods \cite{farsiu2004fast, hardie1997joint}, interpolation \cite{dai2007soft, hou1978cubic, sun2008image} and sparse representation learning \cite{yang2008image, JianchaoYang2010}. More recently, DL approaches have set the new state-of-the-art in SISR regarding image reconstruction accuracy, inference speed, and model generalization, thereby enabling real-time inference at resolution as high as 4K for video applications \cite{kim2018real, dong2016accelerating, wang2018esrgan}. As alluded to above, our work focuses on adapting a traditional SISR method to the quantum setting; thus, in this paper, we skip a detailed discussion on state-of-the-art DL-based approaches. Interested readers may refer to  \cite{bashir2021comprehensive, chen2022real} for a comprehensive overview of the recent state-of-the-art in SISR.

\smallskip
\noindent
\textbf{\textit{(b)} Quantum computing and its current state in computer vision.}
At present, central processing units (CPUs) and quantum processing units (QPUs) are used concurrently, forming a hybrid quantum-classical algorithm. The advantage offered by quantum computing comes at one or more particular critical steps in the algorithm design consisting of problems that are conventionally expensive to solve with only CPUs, such as integer factorization \cite{jiang2018quantum}, graph cut \cite{doan2022hybrid, tse2018graph}, unstructured search \cite{jiang2017near}, and other important combinatorial optimization problems \cite{mcgeoch2013experimental}. To that end, applying quantum computing to such problems can result in commendable speedup over CPU computing by exploiting quantum parallelism \ie, the ability to perform operations on exponentially many superimposed memory states simultaneously \cite{rieffel2000introduction}.

Within modern quantum computing, two paradigms exist to solve different problems amenable to quantum parallelism: universal gate quantum computing and adiabatic quantum computing (AQC). In this paper, we focus on the adiabatic quantum computing paradigm. AQC is a computational method that uses the quantum annealing process to solve a quadratic unconstrained binary optimization (QUBO) problem of the form

\begin{equation}\label{rw_qubo}
\begin{aligned}
\operatorname*{argmin}_\mathbf{z} \mathbf{z}^{T}\mathbf{Q}\mathbf{z}+\mathbf{b}^{T}\mathbf{z},
\end{aligned}
\end{equation}
where, $\mathbf{Q} \in \mathbb{R}^{N \times N}$ contains quadratic QUBO coefficients, $\mathbf{b} \in \mathbb{R}^{N}$ contains linear QUBO coefficients, and $\mathbf{z} \in \{ 0,1 \}^{N}$ contains the binary variables for optimization.

As QUBO is a combinatorial optimization problem that AQC can solve, formulating a computer vision problem as a hybrid classical-quantum algorithm with a QUBO step may enable the computation of the problem to be accelerated compared to a purely classical computing algorithm. In the same spirit, several works in computer vision research have used state-of-the-art AQC hardware containing thousands of qubits to demonstrate the application of AQC to problems such as point correspondence \cite{golyanik2020quantum}, shape matching \cite{benkner2021q}, object tracking \cite{zaech2022adiabatic}, and others \cite{Doan_2022_CVPR, Meli_2022_CVPR, Yang_2022_CVPR}. This work proposes a SISR algorithm developed on D-Wave Advantage quantum annealer QPUs available via the D-Wave Leap interface. Contrary to other quantum computer vision applications, our work is an initial attempt to solve SISR where the previous approaches may not be suitable---reasons will become clear in the later part of the paper. 

In the following section, we provide a brief overview of the concepts on which our approach is developed.

\section{Preliminaries}
\paragraph{Sparse Signal Representation.}
The application of quantum computing to SISR requires the development of an algorithm that contains a step with a problem amenable to solving by QPUs. Current state-of-the-art approaches to SISR based on deep neural networks do not readily satisfy this criterion and as such a suitable SISR problem formulation without involving neural networks design is required. One popular and outstanding SISR approach that could be adapted for AQC is sparse signal representation of SISR problem via $l_1$ minimization of the optimization variable in a cost function. To that end, we derive SISR as a sparse $l_1$ quantum optimization problem, much of which is inspired from the work of Yang et al. \cite{yang2008image}. In our formulation, sparse coding is employed as the basic computational problem to be adapted for quantum computing.

In sparse coding, the output $\mathbf{x} \in \mathbb{R}^{M}$ is recovered as a linear combination of  $\boldsymbol\alpha \in \mathbb{R}^{N}$---usually sparse, and a overcomplete dictionary $\mathbf{D} \in \mathbb{R}^{M \times N}$ in the form $\mathbf{x} = \mathbf{D}\boldsymbol\alpha$. Here, $M$ is the problem dimensionality, $N$ is the number of atoms in the dictionary, while dictionary overcompleteness entails $N>M$.
Yet, the use of QPUs to carry out binary optimization results to a binary sparse coding formulation, where $\boldsymbol\alpha$ consists of binary variables \ie, $\boldsymbol\alpha \in \{ 0,1 \}^{N}$ rather than continous floating values. In SISR via sparse representation, the basic take is to train a dictionary pair $\left( \mathbf{D}_{l},\mathbf{D}_{h} \right)$ from image patches and then to use sparse approximation patch-wise to optimize $\boldsymbol\alpha$ w.r.t. $\mathbf{D}_{l}$ and the LR patches, subsequently generating pixel intensity information for each HR patch from $\mathbf{D}_{h}$ and the optimized $\boldsymbol\alpha$.

\section{Proposed Approach}

\subsection{Problem Formulation}\label{section_qsc_problem}
\noindent
SISR via sparse signal representation that uses classical $l_1$ optimization is popularly formulated as

\begin{equation}\label{eqn:pa_problem}
\begin{aligned}
\boldsymbol\alpha^{*} = \operatorname*{argmin}_{\boldsymbol\alpha} \| \mathbf{D}_{l}\boldsymbol\alpha-\tilde{\mathbf{y}}  \|_{2}^{2} + \lambda \| \boldsymbol\alpha \|_{1}.
\end{aligned}
\end{equation}
To find the sparse coefficients for recovery of HR image patches by $\mathbf{x}=\mathbf{D}_{h}\boldsymbol\alpha^{*}$, Eq.\eqref{eqn:pa_problem} is the main optimization problem solved at inference time. Instead of using conventional algorithms such as LARS \cite{efron2004least}, an AQC-based algorithm could be used. To adapt the optimization problem in Eq.\eqref{eqn:pa_problem} to a format that facilitates AQC, a QUBO formulation involving binary vectors to be optimized is necessary. To this end, a binary mask $\mathbf{m}$ is introduced, leading to a new optimization objective with the binary mask $\mathbf{m}$ 
\begin{equation}\label{eqn:l1o_2}
\begin{aligned}
\mathcal{L} &= \| \mathbf{D}_{l} (\mathbf{m}\odot \mu\mathbf{1})-\tilde{\mathbf{y}} \|_{2}^{2} + \lambda \| (\mathbf{m}\odot \mu\mathbf{1}) \|_{1},  \\
\end{aligned}
\end{equation}
%
%Defining $D_{m}=\text{diag}(m)$, i.e., the matrix constructed 
where $\mathbf{1} = [1,1, \ldots , 1]^{T} \in \mathbb{R}^{N}$. In Eq.\eqref{eqn:l1o_2}, in contrast to the conventional $l_1$ optimization of $\boldsymbol\alpha$, to solve the problem with QUBO, $\boldsymbol\alpha$ is set to be $(\mathbf{m}\odot \mu\mathbf{1})$, where $\mathbf{m}$ is optimized while $\mu\mathbf{1}$ is a fixed vector with all elements set to a hyperparameter $\mu > 0$. This leads to an approximate non-negative $l_1$ optimization problem with the loss term simplified as 
\begin{comment}
\begin{equation}\label{eqn:l1o_3}
\begin{aligned}
&\operatorname*{argmin}_m \| \mathbf{D}_{l} (m\odot \mu \mathbf{1})-\tilde{y} \|_{2}^{2} + \lambda \| (m\odot \mu \mathbf{1}) \|_{1} \\
= \ &  \operatorname*{argmin}_m \| \mu \mathbf{D}_{l} m-\tilde{y} \|_{2}^{2} + \lambda \mu \|  m \|_{1},  \\
\end{aligned}
\end{equation}
\end{comment}
%
\begin{equation}\label{eqn:l1o_5}
\begin{aligned}
 %&= \| \mathbf{D}_{l} (m\odot \mu\mathbf{1})-\tilde{y} \|_{2}^{2} + \lambda \| (m\odot \mu\mathbf{1}) \|_{1}  \\
 \mathcal{L}  &= \| \mu \mathbf{D}_{l} \mathbf{m}-\tilde{\mathbf{y}} \|_{2}^{2} + \lambda \mu \|  \mathbf{m} \|_{1}  \\
 &=  \mathbf{m}^{T} \left(\mu^{2} \mathbf{D}_{l}^{T}\mathbf{D}_{l} \right)\mathbf{m} - 2  \left(\mu  \mathbf{D}_{l}^{T}\tilde{\mathbf{y}} \right)^{T}\mathbf{m} +   \lambda \mu  \mathbf{1}  ^{T}\mathbf{m}  + \text{c} \\
&=  \mathbf{m}^{T} \left( \mu \mathbf{D}_{l}^{T}\mathbf{D}_{l} \right)\mathbf{m} + \left( -2   \mathbf{D}_{l}^{T}\tilde{\mathbf{y}}  +   \lambda  \mathbf{1} \right)^{T} \mathbf{m}  + \text{c},\\
\end{aligned}
\end{equation}
where `$\text{c}$' symbolizes constant w.r.t optimization variable. Comparing Eq.\eqref{eqn:l1o_5} with the basic form of the QUBO problem in Eq.\eqref{rw_qubo} \ie, $\mathcal{L}=\mathbf{m}^{T}\mathbf{Q}\mathbf{m} + \mathbf{b}^{T}\mathbf{m}$, the matrix $\mathbf{Q}$ and the vector $\mathbf{b}$ are given by

\begin{equation}\label{eqn:l1o_6}
\begin{aligned}
\mathbf{Q} &=  \mu \mathbf{D}_{l}^{T}\mathbf{D}_{l}  \\
\mathbf{b} &= -2  \mathbf{D}_{l}^{T}\tilde{\mathbf{y}}  +   \lambda \mathbf{1}.   \\
\end{aligned}
\end{equation}

\subsection{Algorithms}\label{sec:pa_algos}
Having defined the basic formulation for SISR in the previous section, three algorithms are outlined---including ours, and tested. With all algorithms being based on the same basic formulation, they differ by the optimization algorithm used for sparse approximation. Hence, they are referred to by their optimization algorithm. Concretely, the optimization algorithms used are \textit{(i)} \textbf{Lasso Regression}, \textit{(ii)} \textbf{Classical Annealing} and \textit{(iii)} \textbf{Quantum Annealing}. The details of each algorithm are discussed in the remaining part of this section.

\paragraph{\textit{(i)} Lasso Regression.}
The overall Lasso Regression algorithm is shown in \textbf{Algorithm} \ref{algo_srsr}. With trained dictionaries $\mathbf{D}_{h}$ and $\mathbf{D}_{l}$, the overall algorithm takes a LR image $Y$ as input and outputs the SR image $X$. Sparse coding is carried out in a patch-wise manner with a $l_1$ optimization problem solved for each patch. To solve the $l_1$ optimization, the $\texttt{scikit-learn}$ Python machine learning library's $\texttt{linear$\_$model.Lasso}$ solver is used, featuring a $\texttt{C}$-optimized coordinate descent algorithm. Upon combining patch-wise estimates into a single SR image estimate, the image is subject to a backprojection procedure with 100 iterations for convergence to a stable solution with respect to a global reconstruction constraint $X^{*} = \text{argmin}_{X} \|X-X_{0}\|  \quad \text{s.t.} \ (HX)\downarrow = Y$, where $H$ is a blurring filter and $\downarrow$ is a downsampling operation.

\begin{algorithm}
\caption{Lasso Regression}\label{algo_srsr}
\begin{algorithmic}
\State \textbf{Input: }\text{Dictionaries $\mathbf{D}_{h}$ and $\mathbf{D}_{l}$, and LR image $Y$}
\State \text{Extract feature map $\tilde{Y}=F(Y)$}
\For{patch $\tilde{\mathbf{y}}_{i}$ in $\tilde{Y}$}
\State $\boldsymbol\alpha^{*} \gets \operatorname*{argmin}_{\boldsymbol\alpha} \| \mathbf{D}_{l}\boldsymbol\alpha-\tilde{\mathbf{y}}_{i}  \|_{2}^{2} + \lambda \| \boldsymbol\alpha \|_{1}$
\State $\mathbf{x}_{i} \gets \mathbf{D}_{h}\boldsymbol\alpha^{*}$
\State $X_{0} \gets \texttt{update}(X_{0},\mathbf{x}_{i})$
\EndFor
\State $X^{*} \gets \operatorname*{argmin}_{X} \|X-X_{0}\|  \quad \text{s.t.} \ (HX)\downarrow = Y$
\State \textbf{Output: }\text{Super-resolution image $X^{*}$}
\end{algorithmic}
\end{algorithm}

\paragraph{\textit{(ii)} Classical Annealing.}
As a step towards adapting SISR via sparse representation to be compatible with AQC, an algorithm based on solving QUBO problems is developed. Adapted from \textbf{Algorithm} \ref{algo_srsr}, the Classical Annealing algorithm (\textbf{Algorithm} \ref{algo_ca}) uses a QUBO-based formulation with the QUBO problem solved conventionally using CPUs via a simulated annealing algorithm from the \texttt{qubovert} library implemented in \texttt{C}. Since in a QUBO problem binary variables are optimized, the binary sparse coding is carried out with the sparse coefficient vector set to be $(\mathbf{m}\odot \mu\mathbf{1})$ as per the formulation in Sec.(\ref{section_qsc_problem}). Following Eq.\eqref{eqn:l1o_6}, to facilitate binary sparse coding, appropriate constructions are made for the problem matrix $\mathbf{Q}$ and bias $\mathbf{b}$ of the QUBO problem.

\begin{algorithm}
\caption{Classical Annealing}\label{algo_ca}
\begin{algorithmic}
\State \textbf{Input: }\text{Dictionaries $\mathbf{D}_{h}$ and $\mathbf{D}_{l}$, and LR image $Y$}
\State \text{Extract feature map $\tilde{Y}=F(Y)$}
\For{patch $\tilde{\mathbf{y}}_{i}$ in $\tilde{Y}$}
\State $\mathbf{Q} \gets  \mu \mathbf{D}_{l}^{T}\mathbf{D}_{l} $
\State $\mathbf{b} \gets -2  \mathbf{D}_{l}^{T}\tilde{\mathbf{y}}_{i}  +   \lambda \mathbf{1}$
\State $\mathbf{m}^{*} \gets \operatorname*{argmin}_{\mathbf{m}} \mathbf{m}^{T} \mathbf{Q} \mathbf{m} + \mathbf{b}^{T} \mathbf{m} $
\State $\boldsymbol\alpha^{*} \gets  \mu \mathbf{m}^{*} $
\State $\mathbf{x}_{i} \gets \mathbf{D}_{h}\boldsymbol\alpha^{*}$
\State $X_{0} \gets \texttt{update}(X_{0},\mathbf{x}_{i})$
\EndFor
\State $X^{*} \gets \operatorname*{argmin}_{X} \|X-X_{0}\|  \quad \text{s.t.} \ (HX)\downarrow = Y$ 
\State \textbf{Output: }\text{Super-resolution image $X^{*}$}
\end{algorithmic}
\end{algorithm}

%\subsubsection*{Quantum Annealing}
\paragraph{\textit{(iii)} Quantum Annealing.}
The Quantum Annealing algorithm (\textbf{Algorithm} \ref{algo_qbsolv}) represents a way for SISR via sparse representation to be solved using QPUs. It is a modification of the \textit{Qbsolv} algorithm developed by D-Wave \cite{booth2017partitioning}. Quantum Annealing uses a $\texttt{CreateQUBO}$ function (\textbf{Algorithm} \ref{algo_qbsolv_createqubo}) to generate a set of QUBO problems to be solved. In \textbf{Algorithm} \ref{algo_qbsolv_createqubo}, the typical sparse coding problem $\operatorname*{argmin}_{\mathbf{m}} \mathbf{m}^{T} \mathbf{A} \mathbf{m} + \mathbf{b}^{T} \mathbf{m}$ is generated patch-wise from the feature-extracted LR image $\tilde{Y}$ and the LR dictionary $\mathbf{D}_{l}$. The sparse coding problem is initially solved using a MST2 multistart tabu search algorithm from the $\texttt{dwave-tabu}$ library. With an initial solution $\mathbf{m}^{*}$ obtained, a heuristic $\texttt{EnergyImpactDecomposer}$ selects the variables of $\mathbf{m}$ that are expected to cause the most change in the energy $E(\mathbf{m})=\mathbf{m}^{T} \mathbf{A} \mathbf{m} + \mathbf{b}^{T} \mathbf{m}$. The selected variables, numbering 32, are kept track of by $\texttt{index}$.

\begin{algorithm}[h]
\caption{Quantum Annealing}\label{algo_qbsolv}
\begin{algorithmic}
\State \textbf{Input: }\text{Dictionaries $\mathbf{D}_{h}$ and $\mathbf{D}_{l}$, and LR image $Y$}
\State $\{Z_{Q}\},\{ E_{Q}\},\{ O_{Q}\} \gets \emptyset$
\State \text{Extract feature map $\tilde{Y}=F(Y)$}
\State $\{ \mathbf{Q} \},\mathbf{M}_{0},\texttt{index} \gets$ \texttt{CreateQUBO($\tilde{Y}$,$\mathbf{D}_{l}$)}
\For{$\mathbf{Q}$ in $\{ \mathbf{Q} \}$}
\State $Z_{Q},E_{Q},O_{Q} \gets$ \texttt{SolveQUBO($\mathbf{Q}$,$N_{reads}$)} 
%\State $\{Z_{Q}\} \gets$ \texttt{append$\left( \{Z_{Q}\},Z_{Q} \right)$}
\State $\{Z_{Q}\} \gets \{ Z_{Q}\} \cup Z_{Q}$
%\State $\{ E_{Q}\} \gets$ \texttt{append$\left(\{ E_{Q}\},E_{Q}\right)$}
\State $\{E_{Q}\} \gets \{ E_{Q}\} \cup E_{Q}$
%\State $\{O_{Q}\} \gets$ \texttt{append$\left(\{ O_{Q}\},O_{Q}\right)$}
\State $\{O_{Q}\} \gets \{ O_{Q}\} \cup O_{Q}$
\EndFor
\For{patch $\tilde{\mathbf{y}}_{i}$ in $\tilde{Y}$}
\State $Z_{i},E_{i},O_{i} \gets$ \texttt{extract$\left(\{Z_{Q}\},\{E_{Q}\},\{O_{Q}\} \right)$} 
\State $\mathbf{m}_{0i} \gets$ \texttt{extract$\left(\mathbf{M}_{0} \right)$} 
\State $\mathbf{x}_{i} \gets 0$
\State $\mathcal{Z}_{i} \gets 0$
\For{read $\mathbf{z}_{ij}$ in $Z_{i}$}
\State $\boldsymbol\alpha^{*} \gets \mu \mathbf{m}_{0i}$
\State $\boldsymbol\alpha^{*}[\texttt{index}_{i}] \gets \mu \mathbf{z}_{ij}$
\State $p_{ij} \gets O_{ij}\exp \left(-\beta E_{ij} \right)$
\State $\mathbf{x}_{i} \gets \mathbf{x}_{i}+p_{ij}\mathbf{D}_{h}\boldsymbol\alpha^{*}$
\State $\mathcal{Z}_{i} \gets \mathcal{Z}_{i}+p_{ij}$
\EndFor
\State $\mathbf{x}_{i} \gets \mathbf{x}_{i}/\mathcal{Z}_{i}$
\State $X_{0} \gets \texttt{update}(X_{0},\mathbf{x}_{i})$
\State $\mathcal{H}_{i} = -\sum_{j}p_{ij}\log(p_{ij})$
\EndFor
\State $X^{*} \gets \operatorname*{argmin}_{X} \|X-X_{0}\|  \quad \text{s.t.} \ (HX)\downarrow = Y$ 
\State \textbf{Output: }\text{Super-resolution image $X^{*}$}
\end{algorithmic}
\end{algorithm}

\begin{algorithm}[h]
\caption{\texttt{CreateQUBO} for Quantum Annealing}\label{algo_qbsolv_createqubo}
\begin{algorithmic}
\State \textbf{Input: }\text{LR image features $\tilde{Y}$ and dictionary $\mathbf{D}_{l}$}
\For{patch $\tilde{\mathbf{y}}_{i}$ in $\tilde{Y}$}
\State $\mathbf{A} \gets  \mu \mathbf{D}_{l}^{T}\mathbf{D}_{l} $
\State $\mathbf{b} \gets -2  \mathbf{D}_{l}^{T}\tilde{\mathbf{y}}_{i}  +   \lambda \mathbf{1}$$  $
\State $\mathbf{m}^{*} \gets \operatorname*{argmin}_{\mathbf{m}} \mathbf{m}^{T} \mathbf{A} \mathbf{m} + \mathbf{b}^{T} \mathbf{m} $ \Comment{(\textcolor{red}{Tabu search})}
%\State $M_{0} \gets$ \texttt{append$\left( M_{0},m^{*} \right)$}
\State $\mathbf{M}_{0}[i N:(i+1) N] \gets \mathbf{m}^{*} $
\State $\texttt{index}_{i} \gets \texttt{EnergyImpactDecomposer} \left( \mathbf{m}^{*},\mathbf{A},\mathbf{b} \right) $
\State $\mathbf{m}_{tmp} \gets \mathbf{m}^{*}$
\State $\mathbf{m}_{tmp}[\texttt{index}_{i}] \gets 0$
\State $\mathbf{d} \gets 2\mathbf{A}[\texttt{index}_{i}]\mathbf{m}_{tmp}$
\State $\mathbf{Q} \gets \mathbf{A}+\text{diag}\left( \mathbf{d}+\mathbf{b} \right)$
%\State $\{Q\} \gets \texttt{AddSubproblem}\left( \{Q\},Q \right)$
\State $\{\mathbf{Q}\} \gets \texttt{AddSubproblem}\left( \{\mathbf{Q}\},\mathbf{Q} \right)$ \Comment{(Figure \ref{fig_qsc_layout2})}
%(+) Explain what is meant by addsubproblem
\EndFor
\State \textbf{Output: }$\{ \mathbf{Q} \},\mathbf{M}_{0},\texttt{index}$
\end{algorithmic}
\end{algorithm}

\begin{figure}[t]
\begin{center}
\includegraphics[width=0.7\columnwidth]{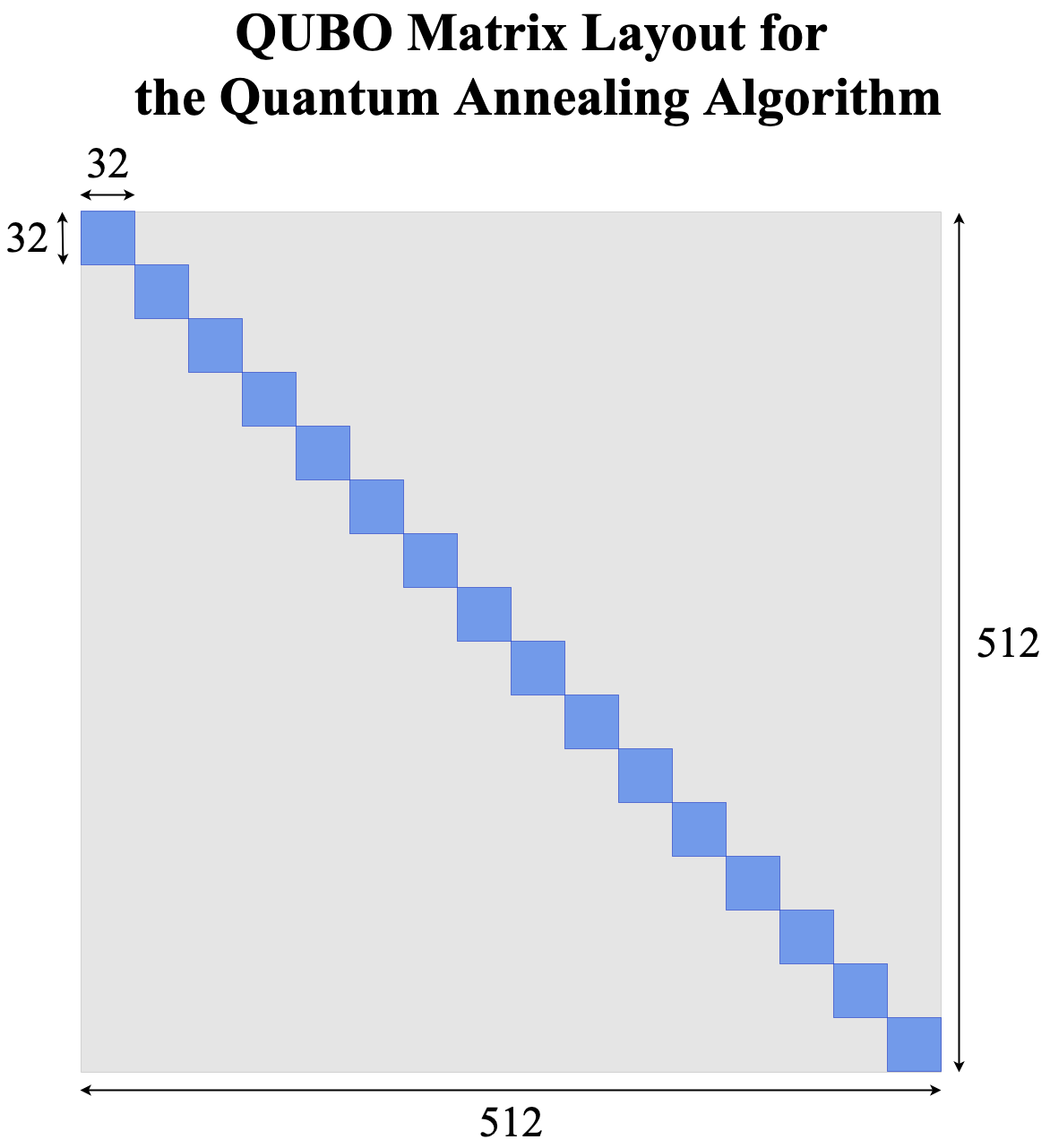}
\end{center}
\caption{QUBO problem matrix layout for Quantum Annealing. The blue parts indicate parts with filled values while the remaining region is not filled with values and as such left out of embedding on QPUs.}
\label{fig_qsc_layout2}
\end{figure}

\noindent
Subsequently, a \textit{clamping} step is applied whereby the non-selected binary variables of $\mathbf{m}$ have their values fixed from the tabu-searched $\mathbf{m}^{*}$ and these values are then used as part of the construction of QUBO subproblems. This leads to the optimization of the 32 highest impact binary variables for each patch. As such, each patch is associated with a QUBO subproblem of size 32 and subproblems are added along diagonals of overall $\mathbf{Q}$ matrices of size 512 as per Fig.\ref{fig_qsc_layout2}. This construction leads to sparse $\mathbf{Q}$ matrices of size $512 \times 512$ that are suitable for the sparse QPU topologies.

The set of QUBO problems constructed is then submitted to a D-Wave direct AQC solver, represented by the $\texttt{SolveQUBO}$ function. Each QUBO problem submitted comes with $N_{reads}$ solutions returned. These solutions are organized in $Z_{Q}$ such that only unique solutions are present. $E_{Q}$ and $O_{Q}$ contain the energy and occurrence of each unique solution. Together, these returned objects organized in the form of sets  $\{Z_{Q}\}$, $\{E_{Q}\}$, $\{O_{Q}\}$ provide a probabilistic distribution over optimal solutions.

Having obtained $\{Z_{Q}\},\{E_{Q}\},\{O_{Q}\}$; $Z_{i},E_{i},O_{i}$ are extracted for each patch $i$ and they are combined with the  tabu-searched initial patch sparse coefficient estimate $\mathbf{m}_{0i}$. The HR $\mathbf{x}_{i}$ is generated by taking the weighted mean over the unique solutions in $Z_{i}$. The weight associated with each unique solution is a \textit{probability} determined by the energy and occurrence as $p_{ij}=(O_{ij}/\mathcal{Z}_{i})\exp \left(-\beta E_{ij} \right)$, where $\mathcal{Z}_{i}= \sum_{j} O_{ij}\exp \left(-\beta E_{ij} \right)$. 
Given the $p_{ij}$ weights, the patch entropy $\mathcal{H}_{i} = -\sum_{j}p_{ij}\log(p_{ij})$ can be defined, providing a metric for uncertainty quantification.

\section{Experiments}

\subsection{Training \& Evaluation Details}
To train the dictionaries $\mathbf{D}_{l}$ and $\mathbf{D}_{h}$, the same procedure and dataset as in Yang et al. \cite{yang2008image} are used. An essential result from Yang et al. \cite{yang2008image} is that dictionaries generated by randomly sampling raw patches from images of similar statistical nature combined with a sparse representation prior are sufficient to generate high quality reconstructions. The training procedure samples raw patches from training images, prunes off patches of small variances and uses an online dictionary learning algorithm \cite{mairal2010online} to generate $\mathbf{D}_{l}$ and $\mathbf{D}_{h}$. In this work, to accommodate the sparse topology of quantum annealers, a relatively small number of atoms $N$ \ie, 128, are learned by $\mathbf{D}_{l}$ and $\mathbf{D}_{h}$. The dataset used to train $\mathbf{D}_{l}$ and $\mathbf{D}_{h}$ in this work is a dataset consisting of 69 images used by Yang et al. \cite{yang2008image}.
In addition to training, hyperparameters for our algorithms are optimized. For Lasso Regression, $\lambda=10^{-5}$ is used. The annealing-based algorithms share a common set of sparse coding hyperparameters with $\lambda=0.1$ and $\mu=0.05$. In Quantum Annealing, $N_{reads}=100$ is used as the number of samples obtained from AQC solvers.

The test data used for evaluation consist of frequently used data in the image processing domain such as the Lenna image and the Set5 image test set. From these data, 3 test cases are set up. The first test case uses a $45\times 60$ cropped region from the Lenna image as the HR ground-truth, with the corresponding downsampled $15\times 20$ region as the LR image (Fig.\ref{fig:lena_eye1}). The second and third test cases are simply the full Lenna image with its downsampled LR version and Set5 respectively.
The first test case of the Lenna image region serves as the basis of a run-time analysis comparing the sparse coding algorithms.
As part of the evaluations, a DL method is compared against the sparse coding algorithms. For this, we used the SwinIR \cite{liang2021swinir}, a current state-of-the-art PSNR-oriented SISR model. The version of SwinIR used is a model pretrained on the DIV2K \cite{agustsson2017ntire} image set with a training patch size of 48.

Due to the varying nature of the algorithms developed, evaluations are naturally carried out on different hardware. Classical algorithms and client-side data processing are carried out on Intel Core 4.00GHz CPUs with 4 cores and 32GB RAM, while the Quantum Annealing algorithm which accesses QPUs directly uses the D-Wave Advantage 6.1 QPU with 5760 qubits.

The accuracy metric used in this work is the Y-channel PSNR between the SR prediction and the HR ground-truth. All downsampling operations reduce image height and width by a factor of 3 and correspondingly all the SR algorithms developed carry out upsampling by a factor of 3.

\begin{figure*}
     \centering
     \begin{subfigure}[b]{0.48\textwidth}
         \centering
         \includegraphics[width=\textwidth]{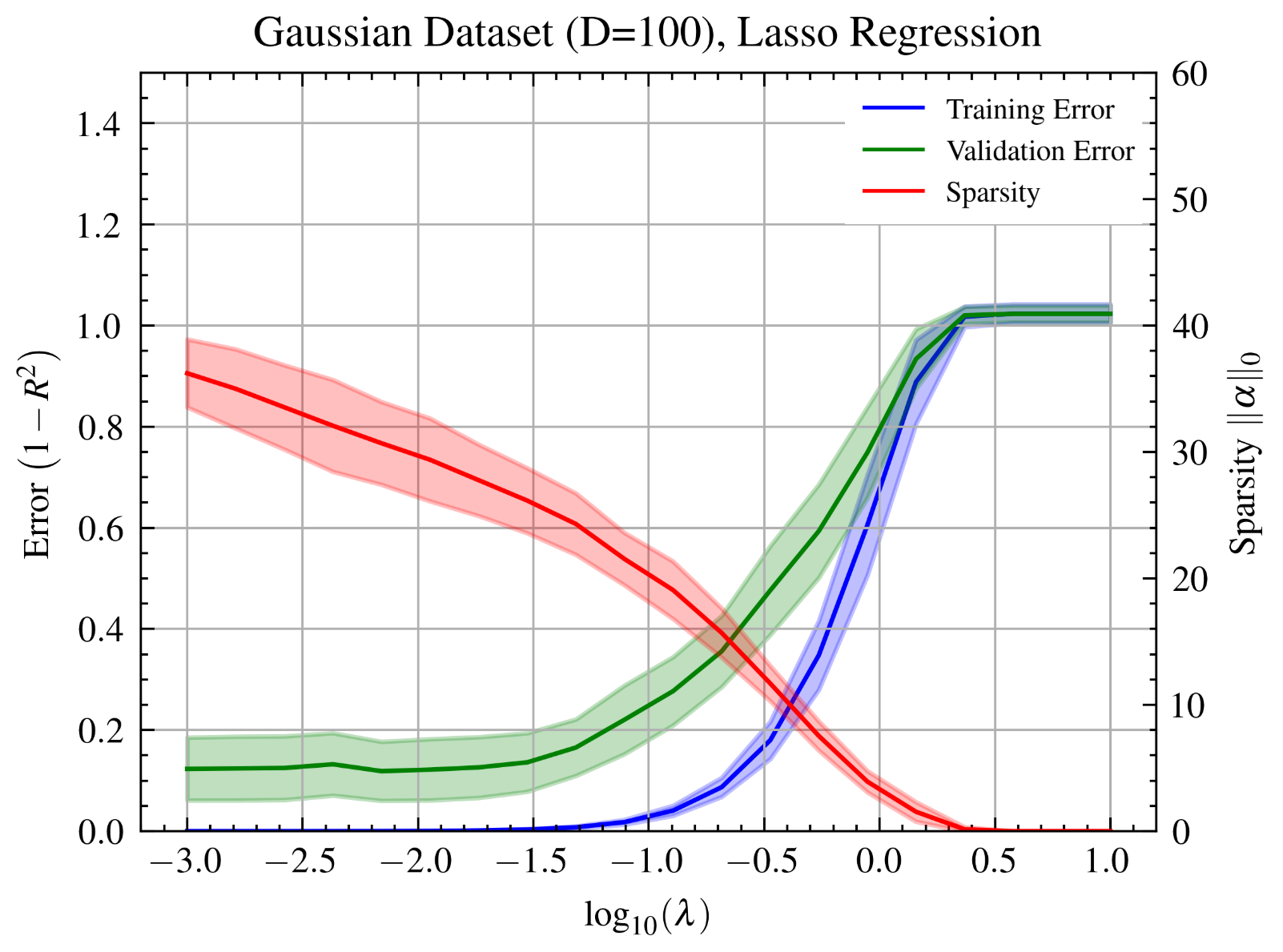}
         \caption{Lasso Regression.}
         \label{fig_num_exp_1}
     \end{subfigure}
     \hfill
     \begin{subfigure}[b]{0.48\textwidth}
         \centering
         \includegraphics[width=\textwidth]{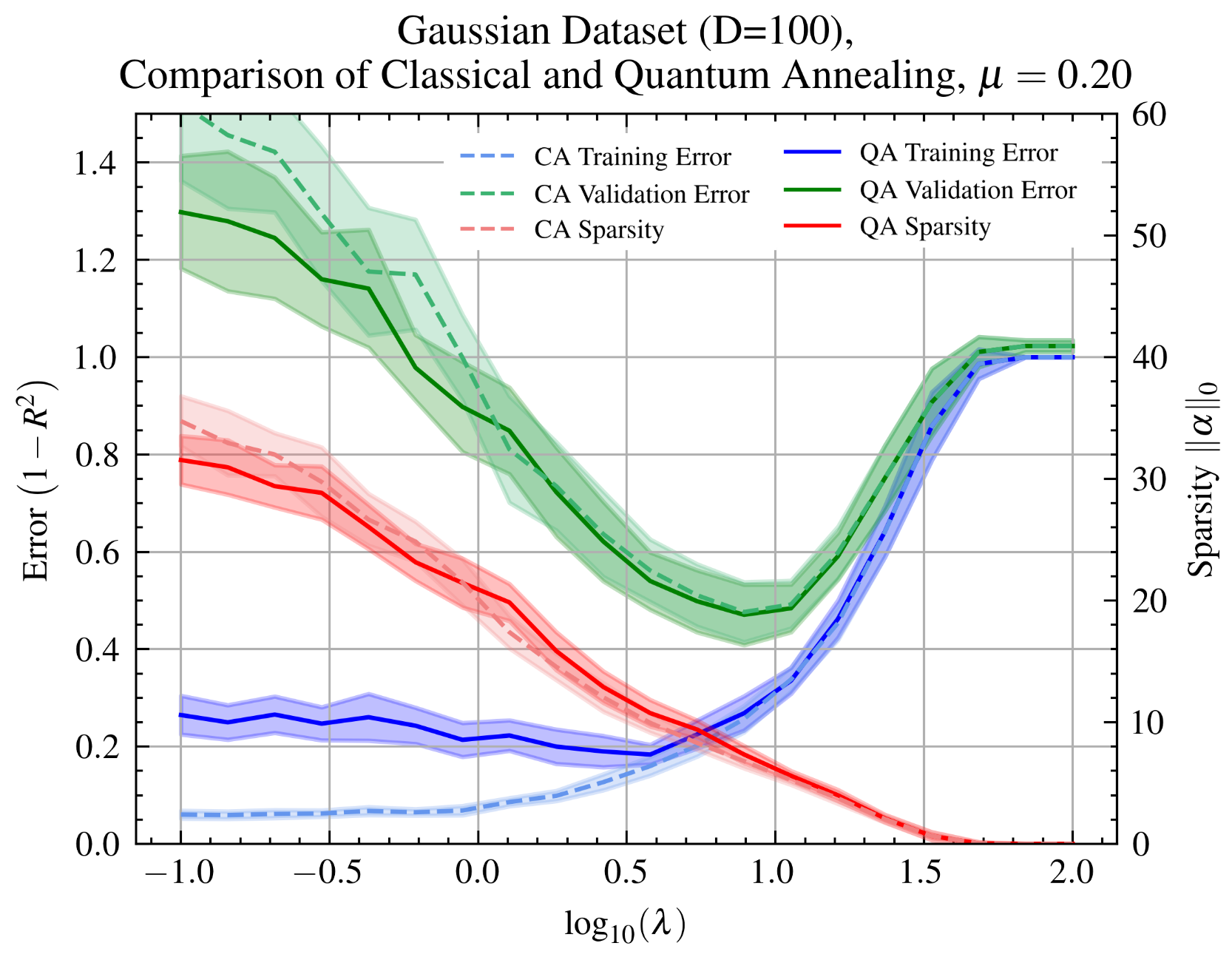}
         \caption{Comparison of Classical Annealing and Quantum Annealing.}
         \label{fig_num_exp_qa}
     \end{subfigure}
        \caption{\textbf{Left:} (a) Numerical results for sparse coding performed by Lasso Regression on 20 datasets generated from a common random process. 
        %The shaded regions show the standard deviations of the relevant quantities. 
        \textbf{Right:} Numerical results for sparse coding performed by Classical Annealing (dashed lines) and Quantum Annealing (solid lines) on 20 datasets generated from a common random process. 
        %The shaded regions show the standard deviations of the relevant quantities. 
        The result for Quantum Annealing is very similar to the result for Classical Annealing especially for the sparsity $\| \boldsymbol\alpha \|_{0} \leq 10$. Note: The shaded regions in the plots show the observed standard deviation.}
        \label{fig_num_exp_overall}
\end{figure*}

\subsection{Numerical Experiments}\label{sec:exp_numexp}
Prior to evaluation on real images, the algorithms developed were first tested on numerical datasets to provide insights about their predictive performance. 
For all numerical experiments detailed in this section, simulated datasets are generated from a common underlying random process. 
Specifically, each dataset is of the form of a data matrix $\mathbf{X}_{all} \in \mathbb{R}^{756 \times 100}$ with all elements sampled from the standard Gaussian distribution $\mathcal{N}(0,1)$. Subsequently, a non-negative random vector $\boldsymbol\alpha_{small} \in \mathbb{R}_{\geq 0}^{10}$ is sampled from $|\mathcal{N}(0,1)|$. $\boldsymbol\alpha_{small}$ is then multiplied with a $\mathbb{R}^{756 \times 10}$ slice of $\mathbf{X}_{all}$ to yield the target vector $\mathbf{y}_{all} \in \mathbb{R}^{756}$. The overall dataset ($\mathbf{X}_{all}$,$\mathbf{y}_{all}$) is partitioned into training and validation sets ($\mathbf{X}_{train}$,$\mathbf{y}_{train}$) and ($\mathbf{X}_{val}$,$\mathbf{y}_{val}$) with sizes 36 and 720 respectively. The resulting setup simulates the sparse coding of a $\mathbb{R}^{36 \times 100}$ dictionary with an optimal sparsity of 10 atoms, as the algorithms tested seek to isolate the 10 features corresponding to $\boldsymbol\alpha_{small}$ from the noisy data.

The numerical results for Lasso Regression and Classical Annealing are shown in Fig.(\ref{fig_num_exp_1}), and the result for Quantum Annealing is shown in Fig.(\ref{fig_num_exp_qa}). In Fig.(\ref{fig_num_exp_1}) a range of values for the sparsity constraint $\lambda$ is evaluated and the prediction error and the $\boldsymbol\alpha$ sparsity level corresponding to each $\lambda$ value are shown. The error metric used for evaluating the predictions $\hat{\mathbf{y}}_{train}$ and $\hat{\mathbf{y}}_{val}$ is $1-R^{2}$ where $R^{2}$ is the coefficient of determination.

Fig.(\ref{fig_num_exp_qa}) shows that Classical Annealing and Quantum Annealing have very similar numerical behavior, no less due to the mathematical similarity of their formulations. For both algorithms, at the known true optimal sparsity level of $\| \boldsymbol\alpha \|_{0}=10$, the validation errors are at around 0.5.
At around $\| \boldsymbol\alpha \|_{0} = 10$, the mean validation error of either annealing-based algorithm lies well within the $1\sigma$ region of the other while there is significant overlap between the $1\sigma$ regions. Thus, there is strong statistical grounding for the similarity of numerical behavior of the two algorithms. In comparison, while Lasso Regression achieves better prediction results at low sparsities, towards the optimal sparsity $\| \boldsymbol\alpha \|_{0}=10$, the validation error converges to those of the annealing-based algorithms. These observations provide evidence that the annealing-based algorithms have similar ability to recover the meaningful features corresponding to $\boldsymbol\alpha_{small}$, hence similar sparse coding performance to Lasso Regression, despite the more limited set of representations achievable by binary sparse coding.

\subsection{Classical \& Quantum Algorithm Comparison}\label{sec:exp_comparison}
The 3 algorithms in Sec.(\ref{sec:pa_algos}) are evaluated and compared against a bicubic interpolation baseline and the SwinIR \cite{liang2021swinir} DL model. The Y-channel PSNRs of these 5 models in total evaluated on the test cases of the Lenna image region, the full Lenna image and the Set5 test set are displayed in Tab.(\ref{table_psnr}).

Tab.(\ref{table_psnr}) shows that for all test cases, Quantum Annealing achieves very similar accuracy as Classical Annealing. In fact, the two algorithms achieve the same PSNR metrics for the larger test cases at a precision of 2 decimal places. This demonstrates that the similar numerical behavior between the two algorithms shown in Fig.(\ref{fig_num_exp_qa}) is replicated when applied to images. Another observation from Fig.(\ref{fig_num_exp_1}) and Fig.(\ref{fig_num_exp_qa}) reflected in Tab.(\ref{table_psnr}) is that the predictive performance of annealing-based algorithms at the very least matches that of Lasso Regression. Going beyond Fig.(\ref{fig_num_exp_1}) and Fig.(\ref{fig_num_exp_qa}), the results for the larger test cases in Tab.(\ref{table_psnr}) furthermore suggest the possibility of superior performance of annealing-based sparse coding over conventional sparse coding when the data is in the form of real images as opposed to simulated data as in Sec.(\ref{sec:exp_numexp}). Nevertheless, the sparse coding approaches have accuracies that fall short of the current state-of-the-art SwinIR DL model.

\begin{table}[ht]
\begin{center}
\begin{tabular}{c | c c c } 
\hline
\scriptsize
Model & \multicolumn{3}{c}{ PSNR }    \\ [0.5ex] 
 &  (a) Lenna reg. & (b) Lenna & (c) Set5 \\ [0.5ex] 
 &  (dB) & (dB) & (dB) \\ [0.5ex] 
 \hline\hline
Bicubic & 28.31 &  30.62 &  29.35\\ 
 \hline
 Lasso Regress. & 30.22  & 31.58 & 30.44 \\ 
 \hline
 Classical Ann. & \second \textcolor{red}{30.20}   & \second \textcolor{red}{31.70} & \second \textcolor{red}{30.61} \\ 
 \hline
Quantum Ann. & \third \textcolor{blue}{30.19}  &  \third \textcolor{blue}{31.70} & \third \textcolor{blue}{30.61} \\ 
 \hline
SwinIR (DL) & \first 31.42 & \first 33.29 & \first 33.89 \\ 
 \hline
\end{tabular}
\end{center}
\caption{PSNR results of the algorithms evaluated on (a) cropped region of the Lenna image (Fig.\ref{fig:lena_eye2}), (b) the full Lenna image Fig.\ref{fig_ca_lena} and (c) the Set5 image set. The magnification factor is 3. The quantum approach generally shows favorable results compared to the classical Lasso. Moreover, our initial primitive quantum experiments for SISR shows encouraging results compared to the state-of-the-art deep learning (DL) method.
}\label{table_psnr}
\end{table}

\begin{figure}[ht]
\begin{center}
\includegraphics[width=1.0\columnwidth]{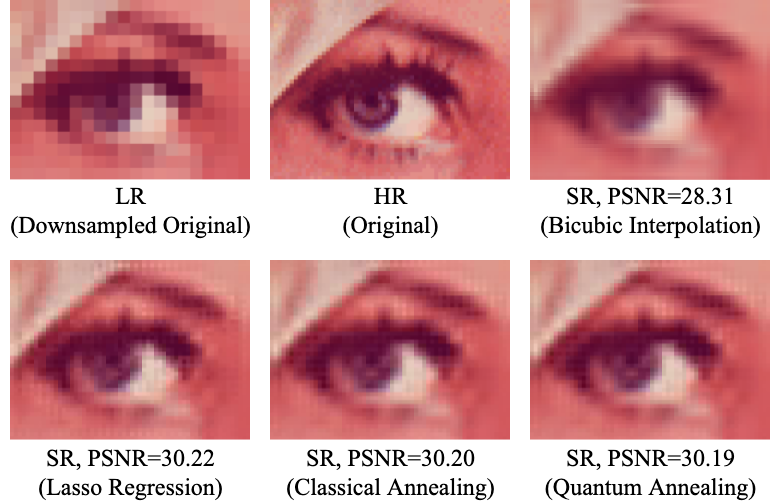}
\end{center}
\caption{Results on a cropped region of the Lenna image with a magnification factor of 3. \textbf{Top row.} \textit{From left to right:} LR, HR and SR using Bicubic interpolation. \textbf{Bottom row.} \textit{From left to right:} SR using Lasso Regression, Classical Annealing and Quantum Annealing respectively.}
\label{fig:lena_eye2}
\end{figure}

\begin{figure}[h]
\begin{center}
\includegraphics[width=1.0\columnwidth]{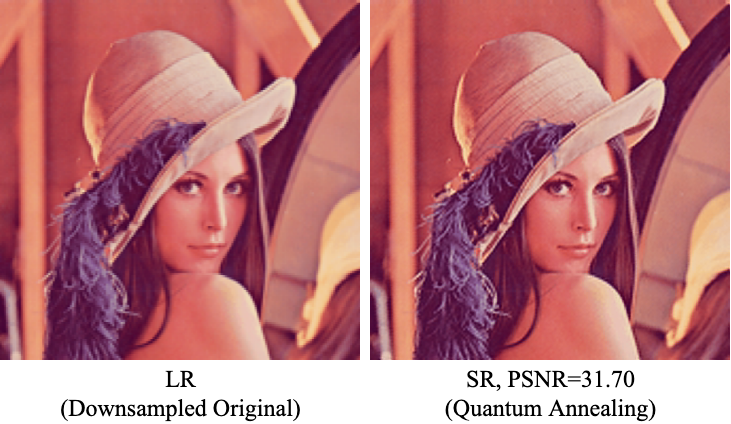}
\end{center}
\caption{ Result on the full Lenna image using the Quantum Annealing algorithm with a magnification factor of 3. \textit{Left:} LR. \textit{Right:} SR using Quantum Annealing. }
\label{fig_ca_lena}
\end{figure}

\begin{table*}[t]
\begin{center}
\begin{tabular}{c | c c c  c c c } 
 \hline
Model & CPU Opt. & CreateQUBO & AQC Prep. &  QPU Opt.  & Misc. & Total Run-time    \\ [0.5ex] 
 & (s) & (s) & (s) & (s) & (s) & (s)  \\ [0.5ex] 
 \hline\hline
Lasso Regression & 4.80 & - & - & -  &  0.09 & 4.89   \\ 
 \hline
 Classical Annealing & 658.24 & -  & - & -  & 0.12  & 658.36   \\ 
 \hline
Quantum Annealing  & - & 31.68& 55.42 & 1.58  &  0.92 & 89.60  \\ 
 \hline
\end{tabular}
\end{center}

\caption{ Run-time results of the algorithms evaluated on a cropped region of the Lenna image (Fig.\ref{fig:lena_eye2}). The total run-time is broken down into 5 constituent run-times, namely CPU Optimization, CreateQUBO, AQC Preparation, QPU Optimization and Miscellaneous. Note that some run-times are not applicable to each algorithm.
%*Overlap=0, **due to job scheduling 
}\label{table_time}
\end{table*}

The run-times of the algorithms on the other hand differ widely, with the annealing-based algorithms requiring more time to run than Lasso Regression. For purposes of analysis, the total run-time of each algorithm is decomposed into 5 run-times. They are namely CPU Optimization, CreateQUBO, AQC Preparation, QPU Optimization and Miscellaneous.

The CreateQUBO run-time refers to the time taken to execute the \texttt{CreateQUBO} function in Quantum Annealing (\textbf{Algorithm} \ref{algo_qbsolv_createqubo}). Executions of \texttt{CreateQUBO} are done exclusively on client-side CPUs. The CPU Optimization run-time refers specifically to the sum of the time taken for all client-side CPU sparse coding steps. As such, it is applicable to executions of the \texttt{scikit-learn} \texttt{linear$\_$model.Lasso} function for Lasso Regression and the \texttt{qubovert.sim.anneal$\_$qubo} function for Classical Annealing, and excludes QPU-based optimization. On the other hand, AQC Preparation and QPU Optimization are run-times applicable to Quantum Annealing. When a QUBO problem is submitted via the D-Wave Leap interface, a period of time is required for communication, scheduling and assignment of the problem to a specific QPU. Subsequently, the QUBO problem is programmed onto a QPU, subject to annealing, sampling and then postprocessed. In this context, the QPU Optimization run-time refers to the combined amount of time taken for programming, annealing and sampling on a QPU, while the AQC Preparation run-time refers to the combined amount of time taken for tasks of communication, scheduling, assignment, postprocessing and other overheads done by the server-side CPU.

Tab.(\ref{table_time}) shows that while Quantum Annealing achieves speed up over Classical Annealing, it is far from matching Lasso Regression's compute time. To improve the speed of Quantum Annealing, one possible approach is to replace the tabu search in \textbf{Algorithm} \ref{algo_qbsolv_createqubo} with a random initialization and omit the use of \texttt{EnergyImpactDecomposer} by instead applying quantum annealing over all subsets of $m$. This approach would drastically simplify the \texttt{CreateQUBO} function but would evidently require more quantum computing resources in terms of access time. The AQC Preparation run-time on the other hand, could not be reduced without fundamental changes to the client-server framework on which this work is based. Nevertheless, should local QPU access be possible, the client-server communication overhead could in principle be eliminated for a much smaller total run-time in which the QPU Optimization run-time becomes the most significant constituent.

\subsection{Robust Prediction \& Uncertainty Estimation}

As discussed in Sec.(\ref{sec:pa_algos}), the D-Wave AQC solver used in Quantum Annealing provides a means to carry out repeated sampling of a QUBO problem to obtain a distribution of solutions, also known as \textit{reads}.
This property provides Quantum Annealing with the advantage of providing a rapid means for robust prediction and uncertainty estimation compared to classical algorithms.

\begin{figure}[h]
\begin{center}
\includegraphics[width=1.0\columnwidth]{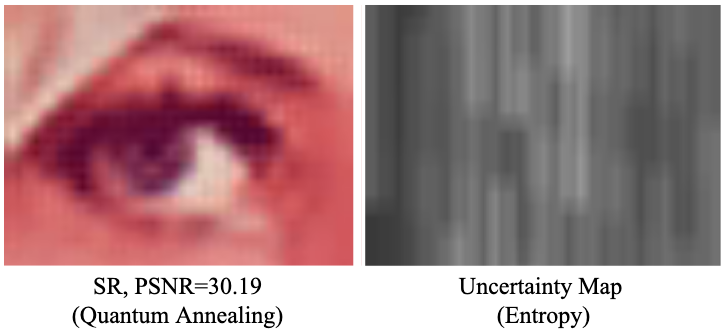}
\end{center}
\caption{ \textit{Left:} Result on a cropped region of the Lenna image (Fig.\ref{fig:lena_eye2}) using the Quantum Annealing algorithm. \textit{Right:} Uncertainty map of the SR image prediction using entropy as the metric for uncertainty. The entropy is computed for each patch $i$ as $\mathcal{H}_{i} = -\sum_{j}p_{ij}\log(p_{ij})$ where $p_{ij}$ is the probability associated with a particular annealing read sample $\mathbf{z}_{ij}$.}
\label{fig_unclena}
\end{figure}

Taking the expectation over a distribution of predictions can give a robust prediction, as done in methods such as Bayesian neural networks \cite{mackay1992practical}. 
$\beta$ effectively determines the number of reads used to generate the final estimate.
%, over which the expectation is taken.
For example in the limit $\beta \rightarrow \infty$, the Quantum Annealing algorithm merely returns a point estimate \ie, the lowest energy solution. On the other hand, when $\beta$ is smaller, more equal weighting is placed on solutions. At a certain optimal $\beta$, the balance between uniqueness and diversity can yield optimally robust estimates. However, due to limited quantum computing resources, insufficient evidence is observed for the advantage offered by such robust prediction from the limited evaluations. As such, further relevant numerical experiments and more extensive image evaluations may be the subject of future work.

Fig.(\ref{fig_unclena}) shows a visualization of the result of uncertainty quantification using Quantum Annealing, where the uncertainty is represented by a map of the entropy. The vertically oriented stripes in Fig.(\ref{fig_unclena}) are due to the column-major nature of the patch-wise iteration in \textbf{Algorithm} \ref{algo_qbsolv}, where neighboring patches along a column are processed in the same QUBO problem.

%conclusion
\section{Conclusion}
Inspired by the early success of sparse coding in computer vision, in this paper, a quantum algorithm for the SISR problem is proposed, discussed, and shown to have a close relationship with the quadratic unconstrained assignment problem---a popular quantum optimization formulation. The summarized quantum annealing algorithm for SISR provides encouraging initial results. While the proposed algorithm is distant from the deep learning-based state-of-the-art image super-resolution result, as an initial demonstration of carrying out SISR with quantum hardware, this work has provided some evidence for superior sparse coding performance of annealing-based methods over conventional sparse coding and insights on how to reduce run-times of the former to comparable levels. Furthermore, the algorithm design, including aspects such as sparse binary coding, QUBO problem layout structuring, and multi-read sampling, can also be applied to the proposed problem settings with advantages offered by quantum annealing, such as superior sparse coding performance, faster robust prediction, and uncertainty estimation.

\noindent
\textbf{Acknowledgement.} The authors thank Jan-Nico Zaech (Ph.D. student at CVL ETH Z\"urich) for useful discussion.

%%%%%%%%% REFERENCES
{\small
\bibliographystyle{ieee_fullname}
\bibliography{egbib}
}

\end{document}